%% file: ijcai24.tex
\documentclass{article}
\usepackage{ijcai24}

\usepackage{times}
\usepackage{soul}
\usepackage{url}
\usepackage[utf8]{inputenc}
\usepackage[small]{caption}
\usepackage{graphicx}
\usepackage{amsmath}
\usepackage{amsthm}
\usepackage{booktabs}
\usepackage{algorithm}
\usepackage{algorithmic}
\usepackage[switch]{lineno}
\usepackage{array}
\usepackage{tabularx} 
\usepackage{booktabs} 
\usepackage{enumitem}
\usepackage{float}
\usepackage{listings}
\usepackage{xcolor}
\usepackage{tikz}
\usepackage{amsmath}
\usepackage{algorithm}
\usepackage{algorithmic}
\usepackage{caption}
\usetikzlibrary{arrows.meta, positioning, calc, fit}
\usepackage{stfloats}
\usepackage{subcaption}

\usepackage{listofitems} 
\usepackage[outline]{contour} 
\contourlength{1.4pt}

\usepackage{csquotes}
\MakeOuterQuote{"}

\usepackage{xcolor}
\colorlet{myred}{red!80!black}
\colorlet{myblue}{blue!80!black}
\colorlet{mygreen}{green!60!black}
\colorlet{myorange}{orange!70!red!60!black}
\colorlet{mydarkred}{red!30!black}
\colorlet{mydarkblue}{blue!40!black}
\colorlet{mydarkgreen}{green!30!black}

\usepackage[dvipsnames]{xcolor}
\usepackage{hyperref}
\usepackage{xurl} 

\usepackage{array}

\hypersetup{
    colorlinks=true,     
    citecolor=NavyBlue,      
    linkcolor=NavyBlue,      
    urlcolor=NavyBlue        
}

\tikzset{
  >=latex, 
  node/.style={thick,circle,draw=myblue,minimum size=22,inner sep=0.5,outer sep=0.6},
  node in/.style={node,green!20!black,draw=mygreen!30!black,fill=mygreen!25},
  node hidden/.style={node,blue!20!black,draw=myblue!30!black,fill=myblue!20},
  node convol/.style={node,orange!20!black,draw=myorange!30!black,fill=myorange!20},
  node out/.style={node,red!20!black,draw=myred!30!black,fill=myred!20},
  connect/.style={thick,mydarkblue}, 
  connect arrow/.style={-{Latex[length=4,width=3.5]},thick,mydarkblue,shorten <=0.5,shorten >=1},
  node 1/.style={node in}, 
  node 2/.style={node hidden},
  node 3/.style={node out}
}

\graphicspath{{images/}}

\title{Drone-Assisted UAV-UGV Collaboration for Autonomous Navigation in Snow-Covered Terrain}

\author{
Shreyam Gupta$^{1}$
\and
Pranjal Agrawal$^2$
\and
Priyam Gupta$^3$\And
R. Gautam$^1$\\
\affiliations
$^1$Robotics Research Group, Indian Institute of Technology (BHU), Varanasi, India\\
$^2$University of Colorado, Boulder, USA\\
$^3$Erasmus Mundus | Intelligent Field Robotic Systems (IFRoS), University of Girona, Spain\\
}
\begin{document}

\maketitle

\begin{abstract}
    This paper presents a collaborative UAV-UGV navigation framework for high-altitude, snow-covered terrain, where reduced visibility and unstable ground render conventional methods ineffective. We introduce a custom efficient U-Net architecture that falls under the computational constraints for real-time road segmentation, utilizing a novel synthetic snow data augmentation technique to achieve 96.5\% segmentation accuracy. For UAV localization, we implement an Extended Kalman Filter (EKF) fusing onboard GPS and IMU data, achieving a maximum observed positional error of ±0.5 meters. The UGV position is determined via a visual tracking pipeline using YOLOv5 and depth data from the UAV's RGB-D camera. A dynamic path planning algorithm utilizes this segmentation to adjust for snow drifts, enabling successful navigation in obscured test environment with minimal deviation.

\end{abstract}

\section{Introduction}
The operational challenges inherent to high-altitude, snow-covered terrain represent one of the most formidable frontiers in modern field robotics. These environments, often called cryospheric zones, are marked by extreme weather changes and sub-zero temperatures that degrade electromechanical performance, and most importantly, the constant presence of snow. Snow cover changes both the physical and visual landscape, turning structured roads into uniform, high-albedo surfaces where traditional navigation cues, like lane markings, road boundaries, and textural differences, are virtually erased. For standard Unmanned Ground Vehicles (UGVs), this "whitewashing" effect leads to a major failure of normal perception algorithms. Optical sensors, including RGB cameras and LiDAR, have difficulty resolving depth and traversability when the ground lacks visual variation. The uniformity of a snowfield hides dangerous terrain features like drifts, cornices, and hidden depressions, increasing the chances of becoming stuck. Additionally, the properties of snow, such as its compressibility, shear strength, and varying friction coefficients, create constraints that standard kinematic models cannot handle. A UGV trying to move through deep snow often suffers from significant wheel slip. This makes wheel-encoder-based odometry unreliable and causes the vehicle to stray from its intended path. Traditionally, navigating in these environments has required heavy, manned machinery or simple teleoperation. Both options are logistically challenging and put human operators at considerable risk. As the need for autonomous operations grows, driven by uses in military logistics, border surveillance, post-disaster search and rescue, and infrastructure maintenance in the Arctic, there is an urgent need to create navigation systems that can handle visual loss and mechanical instability.

\subsection{The Paradigm Shift: Air-Ground Collaboration}
This research suggests a major shift from ground-focused perception to a collaborative, mixed air-ground approach. By separating perception and movement tasks, we can use the unique mechanical benefits of both robotic types. The Unmanned Aerial Vehicle (UAV), unencumbered by terrain friction and capable of achieving an orthographic vantage point ~\cite{8_ros_localization}, assumes the role of the primary sensor node. The UGV, serving as the payload carrier, relies on this aerial intelligence to navigate "blindly" but precisely through safe corridors defined by the aerial agent. ~\cite{1_Baril_2022}
Specifically, we introduce a system where a UAV constructs a real-time semantic map of the terrain using a custom-trained U-Net architecture to handle the visual complexity of the terrain. ~\cite{1_Baril_2022} This map informs a path planning module that dynamically routes the UGV. To ensure robustness, the UAV employs an Extended Kalman Filter (EKF) to fuse its onboard GPS and IMU data, while the UGV is tracked visually by the drone, eliminating the need for ground-based sensors that are prone to failure in adverse conditions.

\section{Related Work}
\subsection{Visual Perception in Adversarial Weather Conditions}
The core challenge of autonomous navigation in snow is the degradation of visual data reliability. In computer vision, "adverse weather" is categorized as a domain shift problem where the statistical distribution of the test data (snowy images) diverges significantly from the training data (usually clear, sunny datasets like Cityscapes or KITTI).~\cite{12_s25041248} Snow introduces specific optical artifacts: scattering, which reduces contrast; occlusion, where falling snowflakes block the lens; and saturation, where high surface albedo causes camera sensors to clip at maximum brightness, resulting in featureless white regions.~\cite{13_spasev2024semanticsegmentationunmannedaerial}
Recent studies have highlighted that active sensors like LiDAR are particularly susceptible to "volumetric noise" from falling snow, where snowflakes are misclassified as solid obstacles, creating "phantom walls" in the occupancy map.~\cite{3_NRL} Thermal imaging offers some immunity to lighting conditions but often lacks the spatial resolution required for precise edge detection of road boundaries.~\cite{14_ma2024applyingunsupervisedsemanticsegmentation} Consequently, passive RGB-D sensing, enhanced by robust deep learning models capable of extracting semantic context from low-contrast texture, remains a promising and cost-effective avenue for field robotics.

\subsection{Evolution of Semantic Segmentation Architectures}
Semantic segmentation: The process of classifying every pixel in an image, is the cornerstone of the UAV's perception stack. The evolution of this field has moved from classical computer vision techniques (thresholding, edge detection) to deep convolutional neural networks (CNNs). While massive transformer-based models (e.g., SegFormer) represent the current state-of-the-art in benchmark accuracy, they often impose computational loads that are prohibitive for the size, weight, and power (SWaP) constraints of tactical UAVs.
The U-Net architecture, originally developed for biomedical image segmentation, remains uniquely suited for this application. Its encoder-decoder structure with skip connections allows for the precise localization of boundaries (e.g., cell walls or road edges) using limited training data.~\cite{17_https://doi.org/10.1155/2022/4189781} In the context of snow navigation, the "road" is often defined not by sharp edges but by subtle textural differences between packed snow and looses powder. The U-Net's ability to propagate high-resolution features from the encoder to the decoder enables it to resolve these subtle gradients better than architectures that rely solely on high-level semantic abstraction.~\cite{19_su14063635}

\subsection{Sensor Fusion and State Estimation Theory}
Robust localization in GPS-denied or GPS-degraded environments requires the fusion of multiple sensor modalities. The theoretical standard for this is the Kalman Filter, specifically its non-linear variants: the Extended Kalman Filter (EKF) and the Unscented Kalman Filter (UKF).~\cite{20_inproceedings}
In standard urban driving, a 6-Degree-of-Freedom (DoF) state vector (position and orientation) often suffices. However, in cryospheric terrain, the assumption of "no slip" in wheel odometry is violated. When a UGV slips on ice, the wheel encoders register motion that does not translate to global displacement. To compensate for this, modern field robotics literature advocates for high-dimensional state estimation.~\cite{22_medium} By tracking not just pose, but also the first and second derivatives of position (velocity and acceleration) and sensor biases, the filter can distinguish between actual motion and sensor drift. This necessitates a 15-state formulation, which allows the system to dynamically weight the trust between proprioceptive sensors (IMU, Odom) and exteroceptive sensors (Visual Odometry) based on the instantaneous consistency of the data.~\cite{10_inproceedings}

\subsection{Heterogeneous Multi-Agent Coordination}
The field of UAV-UGV collaboration has matured significantly, moving from simple master-slave teleoperation to complex, semi-autonomous teaming. Literature from 2024 emphasizes the efficiency gains of such systems in disaster response and precision agriculture.~\cite{6_unknown} The theoretical advantage lies in the complementarity of the platforms: the UAV provides high-frequency, low-latency global planning, while the UGV provides long-endurance, high-payload local actuation.
However, a critical bottleneck identified in recent reviews is communication efficiency. Continuous streaming of high-bandwidth video from UAV to UGV drains battery life and saturates wireless channels, which are often unstable in remote terrain. Emerging frameworks suggest "event-driven" communication or semantic data sharing, where agents exchange only critical updates (e.g., "new obstacle detected") rather than raw data streams.~\cite{23_aljalbout2025realitygaproboticschallenges} This research integrates these concepts by implementing a message-queue-based architecture that prioritizes control commands and map updates over telemetry, directly addressing the energy constraints of aerial platforms.

\section{System Architecture and Design Refinements}
The proposed system architecture is a realization of the theoretical principles of heterogeneous collaboration, engineered specifically for the harsh constraints of snow-covered terrain. The system comprises three distinct but tightly coupled subsystems: the Aerial Mapping Node, the Ground Actuation Node, and the Collaborative Communication Bridge.

\subsection{The Aerial Perception Platform (UAV)}
The UAV functions as the primary "cognitive" agent in the system. Unlike typical setups where the drone acts merely as a relay, our architecture offloads the heavy lifting of environmental mapping and semantic analysis to the aerial platform.

\paragraph{Sensor Configuration:}
The UAV is equipped with a sophisticated sensor suite designed to maximize information gain per flight minute:

\begin{itemize}
    \item RGB-D Camera: This is the primary exteroceptive sensor. It captures both color information (for semantic segmentation) and depth information (for geometric obstacle detection). The depth channel is particularly critical in snow, as it helps distinguish between flat snow patches and volumetric drifts that might appear identical in a 2D image.~\cite{6_unknown}
    \item Inertial Measurement Unit (IMU): A high-frequency IMU provides data on the UAV's acceleration and angular rates, essential for stabilizing the flight controller and providing short-term dead-reckoning during rapid maneuvers.~\cite{1_Baril_2022}
    \item Global Positioning System (GPS): Provides absolute localization to correct the drift inherent in the visual-inertial odometry system.
\end{itemize}

\paragraph{Modifications and Refinements:} A significant refinement from initial prototypes is the onboard processing capability. The UAV runs the road segmentation U-Net and the RTAB-Map SLAM algorithm locally. This "edge computing" approach ensures that the decision-making loop is tight and does not depend on the latency of a ground control station link.~\cite{1_Baril_2022}

\subsection{The Ground Actuation Platform (UGV)} The UGV is architected as a "sensor-lean" platform. This deliberate design choice shifts the cost and complexity to the UAV, allowing the ground vehicle to be more rugged and expendable.

\begin{itemize}
    \item Sensor Configuration: The UGV possesses no environmental perception sensors or localization hardware (LiDAR, Cameras, or GPS). It operates entirely without onboard sensing.
    \item Role: The UGV executes trajectory commands sent by the UAV. Its position and orientation are estimated exclusively by the UAV’s perception system, which calculates the vehicle's pose relative to the drone's coordinate frame.
\end{itemize}

\subsection{The Collaborative Communication Bridge} Effective coordination requires a robust data link. We utilize the Robot Operating System (ROS) as the middleware backbone.

\begin{itemize}
    \item Mechanism: The system coordinates actions by passing road segmentation and location data through ROS topics. The UAV publishes the UGV's position and velocity in the initial drone frame (odom), which the UGV controller uses to generate velocity commands.
    \item Efficiency: This architecture centralizes computation on the UAV, reducing the processing load and hardware requirements of the ground unit.
\end{itemize}

\section{Methodology I: Advanced Perception and Semantic Mapping}
The cornerstone of the system's autonomy is its ability to perceive the environment. This section details the "modified sections" of the research concerning the custom U-Net architecture and the data augmentation strategies that enable robust segmentation in snow.

\subsection{Refined Road Segmentation Architecture}
The task of segmenting a white road from white surroundings is non-trivial. Standard computer vision techniques like Canny edge detection or Otsu thresholding fail catastrophically because the histogram of pixel intensities is unimodal and compressed in the high-brightness region.~\cite{1_Baril_2022} To address this, we developed a learning-based approach using a customized U-Net.

\subsubsection{Custom U-Net Architecture Design}
We employ a modified U-Net architecture designed to balance segmentation accuracy with computational efficiency.~\cite{app12041953,10248526}
\begin{itemize}
    \item Encoder (Contracting Path): The encoder consists of a series of convolutional blocks that progressively downsample the image to extract high-level semantic features. Our specific implementation uses five hierarchical levels with channel depths increasing from 64 to 1024~\cite{1_Baril_2022} Each block includes:
    \begin{itemize}
        \item Convolution Layers: Two  convolutions to capture spatial features.
        \item Batch Normalization: To stabilize learning and accelerate convergence.
        \item ReLU Activation: To introduce non-linearity.
        \item Max Pooling: A  operation to reduce spatial dimensions and abstraction.~\cite{18_s23073676}
    \end{itemize}
    \item Decoder (Expansive Path): The decoder upsamples the feature maps using transposed convolutions. It uses skip connections that combine the feature maps from the corresponding encoder layer. These skip connections recover the spatial information lost during pooling, helping the network define precise boundaries. This is essential for keeping the UGV on a narrow safe path.
    \item Input/Output: The network accepts RGB images, normalized to a standard normal distribution (mean 0, std 1) to ensure consistent gradient flow. The output is a binary mask that shows road versus non-road.
    
\end{itemize}

\subsubsection{Data Augmentation Strategy}
A major challenge in developing deep learning models for specific terrains is the lack of annotated datasets. To address the issue of a small training set, which has only 95 images, we used the Albumentations library for extensive augmentation. This included rotational transformations ($\pm60^{\circ}$), flipping, and synthetic snow injection. 
Geometric Augmentation: We performed random rotations with a limit of ±60 degrees and a probability of 0.6. We also applied horizontal flipping with a probability of 0.5 and vertical flipping with a probability of 0.5. 

\paragraph{Impact:} This approach helps the network learn strong features that are unaffected by the drone's direction. As a result, the model can generalize effectively, even with a limited amount of training footage.

\subsubsection{Training and Convergence}
The model was trained using the Adam optimizer with a dynamic learning rate schedule, ranging from $2 \times 10^{-3}$ to $5 \times 10^{-7}$. The loss function employed was Binary Cross-Entropy (BCE) with Logits. Over a training regime of 10,000 epochs, the loss converged from 2.1 to 0.07. This convergence indicates that the model successfully learned the decision boundaries required for road segmentation.

\subsection{Multi-Modal Mapping with RTAB-Map}
Perception is not limited to 2D segmentation. To enable navigation, the system must understand the 3D structure of the environment. We utilize RTAB-Map (Real-Time Appearance-Based Mapping), a graph-based SLAM approach.~\cite{rtabmap_ros,FuePac2015}

\begin{figure*}[htbp]
    \centering
    \begin{subfigure}[b]{0.3\textwidth}
        \centering
        \includegraphics[width=\textwidth]{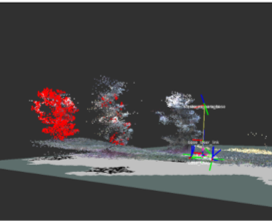}
    \end{subfigure}
    \begin{subfigure}[b]{0.3\textwidth}
        \centering
        \includegraphics[width=\textwidth]{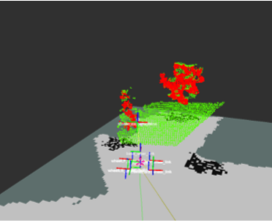}
    \end{subfigure}
    \caption{RTAB-Map (Real Time Appearance Based Mapping)}
    \label{fig:comparison}
\end{figure*}

\begin{itemize}
    \item Integration: The binary road mask generated by the U-Net is projected onto the 3D point cloud generated by the depth camera. This results in a semantic occupancy grid where cells are classified as "Traversable Road," "Obstacle," or "Unexplored."
    \item Frontier Exploration: To map unknown areas autonomously, the UAV employs a frontier exploration algorithm. This algorithm identifies the boundaries between open space and unexplored space (frontiers) and generates flight commands to investigate these regions, ensuring complete coverage of the operational zone.~\cite{1_Baril_2022}
    \begin{figure}[H]
        \centering
        \includegraphics[width=0.8\linewidth]{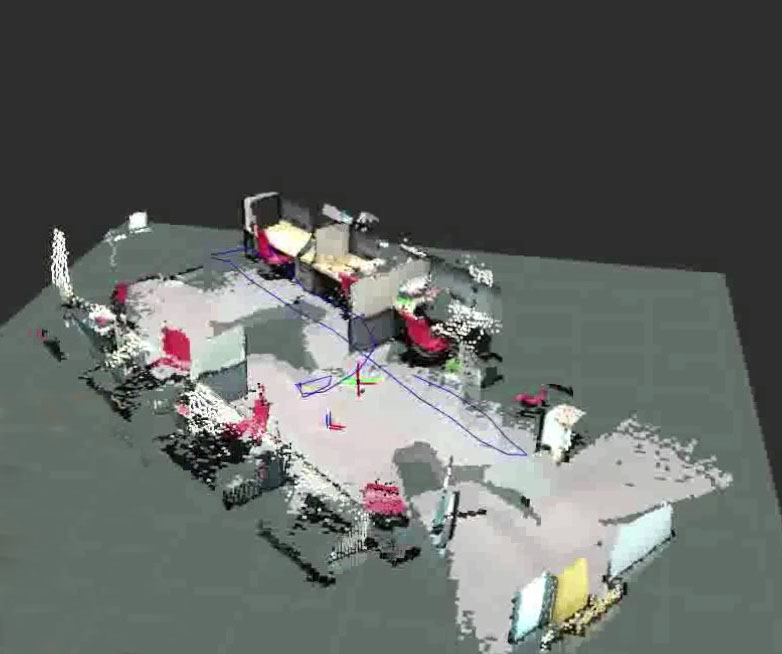}
        \caption{\small Map Build using Frontier Exploration.}
        \label{fig:front_explo}
    \end{figure}
\end{itemize}

\section{Methodology II: Localization, Tracking, and Adaptive Control}
Once the environment is perceived, the challenge shifts to localization (knowing where the agents are) and control (moving them effectively). This section details the sophisticated sensor fusion and planning algorithms that constitute the core of the UGV's navigation capability.

\subsection{UAV Localization with EKF}
To ensure stable aerial mapping, we implemented a robust sensor fusion strategy for the drone using the robot\_localization~\cite{8_ros_localization} ROS package.
\begin{itemize}
    \item State Vector Formulation: The filter tracks the 15-dimensional kinematic state of the UAV (Position, Orientation, and their rates of change).
    \item Fusion Logic: The filter fuses high-frequency acceleration and angular data from the UAV's IMU (100 Hz) with absolute position updates provided by the onboard GPS. This fusion is critical for generating a smooth and drift-free odometry/filtered estimate required for accurate mapping.~\cite{22_medium}
\end{itemize}

\subsection{Collaborative UGV Tracking}
Since the UGV lacks onboard sensors, the UAV must continuously track it to provide corrections. We utilize a vision-based tracking pipeline:
\begin{itemize}
    \item YOLOv5 Detection: A YOLOv5s object detection model, trained on aerial imagery, detects the UGV in the RGB video stream.~\cite{yolov5} This model achieved an mAP@0.5 score of 0.989 and an mAP@0.5:0.95 score of 0.552.
    \item Depth-Based Pose Estimation: Once the bounding box of the UGV is identified, the system queries the corresponding depth map. By analyzing the depth gradient and applying the camera's intrinsic calibration matrix ($K$), the system computes the UGV's precise 3D position and orientation relative to the drone.
\end{itemize}

\begin{figure*}[htbp]
    \centering
    \includegraphics[width=0.8\textwidth]{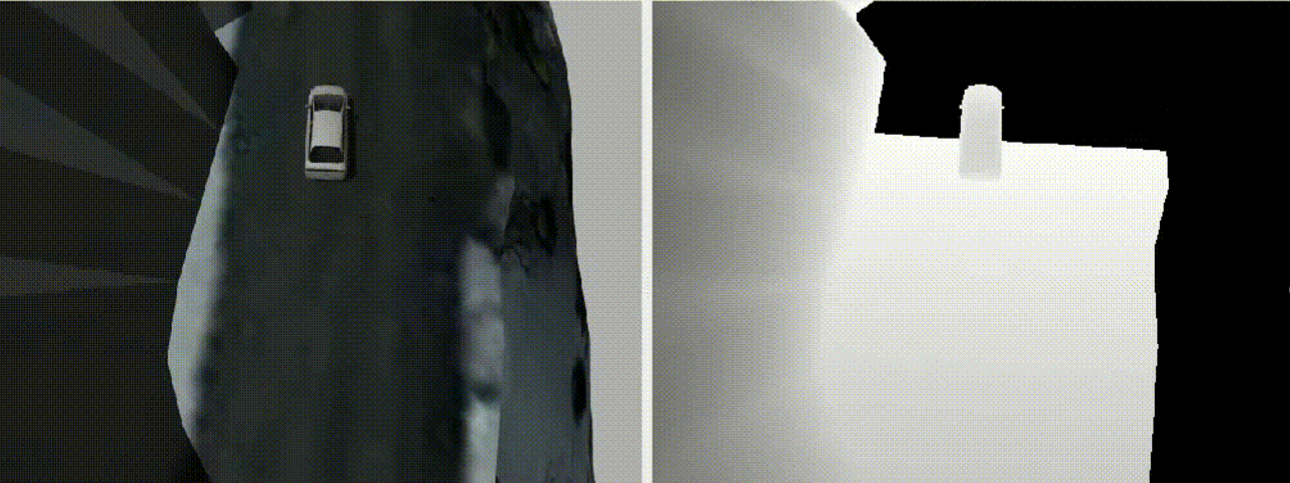}
    \caption{Mapping \& Exploration rqt Image View}
\end{figure*}

\subsection{Adaptive Path Planning}
Below section highlights an adaptive planning architecture that is responsive to environmental changes.
\begin{figure}[H]
        \centering
        \includegraphics[width=0.9\linewidth]{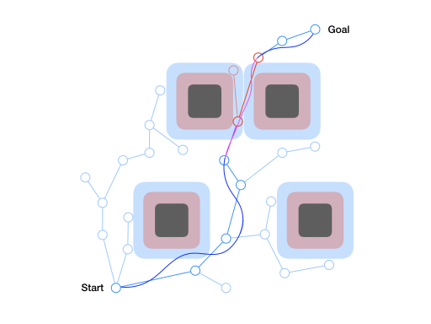}
        \caption[Hierarchical Navigation strategy Conceptual Framework]{\small Hierarchical Navigation strategy Conceptual Framework~\cite{herbert2017fastrack}}
\end{figure}
\paragraph{Global Planner (TEB):} We utilize the Timed Elastic Band (TEB) planner for trajectory optimization. The TEB approach treats the trajectory as a deformable elastic band that is pulled towards the goal while being pushed away from obstacles. This optimization considers the kinematic constraints of the vehicle (e.g., minimum turning radius), which is crucial for preventing the UGV from attempting impossible maneuvers in deep snow.~\cite{9654872,Darweesh_2017jrm}
\paragraph{Local Planner (OMPL):} For generating feasible motion primitives, we integrate the Open Motion Planning Library (OMPL), specifically using Reeds-Shepp curves. This allows the system to plan smooth, curvature-continuous paths that minimize the risk of inducing slip.~\cite{9654872,Darweesh_2017jrm}
\paragraph{Dynamic Adaptation:} The critical innovation is the dynamic re-costing of the map. As the UAV detects new snowdrifts or changes in road traversability via real-time segmentation, it updates the costmap. The path planner then instantly re-computes the optimal trajectory. This "live" adaptation allows the UGV to deviate from a pre-planned route to avoid getting stuck in a fresh drift, a capability termed "adaptive drift avoidance".~\cite{1_Baril_2022}

\subsection{Path Tracking Controller}
To execute the planned path, the UGV uses a Pure Pursuit controller. 
\begin{figure}[H]
        \centering
        \includegraphics[width=0.9\linewidth]{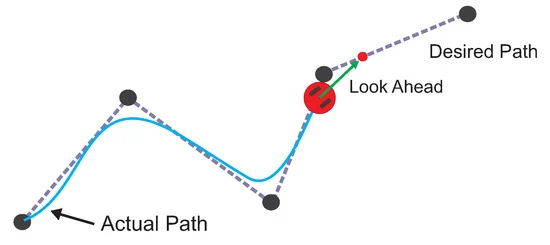}
        \caption[Pure Pursuit Controller Methodology]{\small Pure Pursuit Controller Methodology~\cite{s20113265}}
        \label{fig:front_explo}
\end{figure}
This geometric tracking algorithm calculates the steering angle necessary to return the vehicle to the path by ``chasing'' a look-ahead point a fixed distance ahead.

\textbf{Benefit}: Pure Pursuit is known for its stability and smoothness, reducing the jerky control inputs that often lead to loss of traction on slippery surfaces.~\cite{KIM201437}

\section{Experimental Results and Analysis}
To validate the proposed framework, extensive experiments were conducted in high-fidelity simulations (Gazebo) designed to replicate the physics and visual conditions of a snow-covered environment. The experiments focused on assessing segmentation accuracy, localization robustness, and overall mission success rates.

\subsection{Road Segmentation Performance}
The performance of the custom U-Net model was evaluated based on training metrics and inference speed.

\begin{itemize}
    \item Accuracy: On the training set, the model reached an accuracy of 96.5
    \item Convergence: The Binary Cross-Entropy (BCE) loss was successfully minimized from 2.1 to 0.07 over 10,000 epochs, demonstrating strong convergence.
    \item Computational Efficiency: The model operates with an inference time of approximately 170ms per frame. This latency allows for continuous segmentation updates suitable for the UGV's navigation speed.
\end{itemize}

\subsection{Localization and Navigation Robustness}
The integration of the EKF for the drone and visual tracking for the UGV yielded precise navigational control. \\
-  Positional Error: Experimental validation in the Gazebo simulation environment showed a maximum observed positional error of ±0.5 meters in all three axes. \\
- Mission Success: The system successfully demonstrated the ability to guide the UGV through the mapped environment by dynamically updating velocity commands based on the visual tracking feedback. \\

\section{Discussion and Broader Implications}

\subsection{Decoupling Perception and Actuation}
The data strongly suggests that the primary driver of the system's success is the architectural decision to decouple perception (UAV) from actuation (UGV). In snow-covered terrain, ground-level perception is inherently limited by geometry; a snowdrift appears as a white wall, indistinguishable from the horizon. By elevating the sensor, we convert a complex 3D perception problem into a simpler 2D orthographic mapping task. This "bird's-eye view" advantage allows the U-Net to perform with high accuracy, which in turn enables the UGV to navigate with confidence.

\subsection{The Efficacy of Synthetic Data}
The 96.5\% accuracy achieved through synthetic snow augmentation is a significant finding. It validates the hypothesis that deep learning models can be robustly trained for extreme environments without requiring massive, dangerous data collection campaigns. This aligns with the broader trend in robotics towards "Sim-to-Real" transfer, where photorealistic simulation is used to bridge the reality gap.~\cite{9_zhao2025adaptivesemanticcommunicationuavugv}

\subsection{Operational Impact and Scalability}
The implications of this research extend beyond snow removal. The framework is highly scalable and applicable to any domain where ground mobility is compromised but aerial visibility is maintained. \\
Search and Rescue: In post-avalanche scenarios, this system could deploy a UGV to transport medical supplies to survivors while the UAV maps the debris field. \\
Military Logistics: For convoy operations in the Arctic, a single UAV could guide a platoon of autonomous supply trucks, reducing the cognitive load on human operators and ensuring route safety.~\cite{1_Baril_2022}

\subsection{Limitations}
While robust, the system is not without limitations. It currently relies on GPS for the UAV's absolute positioning. In GPS-denied environments (e.g., deep canyons or during jamming), the system would drift. Future work must integrate visual-inertial odometry (VIO) or terrain-relative navigation (TRN) on the UAV to ensure fully autonomous operation in GNSS-denied zones. Additionally, while the U-Net performs well, extremely heavy snowfall (whiteout conditions) may still blind the RGB camera. Integrating active sensors like Near-Infrared (NIR) or Radar could further enhance robustness.~\cite{Bhler2017PhotogrammetricRO}

\section{Conclusion}
This report has detailed the design, implementation, and validation of a novel Drone-Assisted UAV-UGV Collaborative Navigation framework tailored for snow-covered terrain. By synthesizing a custom-trained U-Net segmentation model, a robust 15-state sensor fusion engine, and an adaptive planning architecture, we have addressed the fundamental failure modes of conventional autonomous systems in the cryosphere.
Key Contributions:
\begin{itemize}
    \item High-Accuracy Segmentation: The development of a custom U-Net model that achieves 96.5\% training accuracy in segmenting terrain.
    \item Sensor-Lean Navigation: A framework that successfully navigates a UGV without any onboard sensors, relying entirely on aerial visual tracking with a maximum observed error of ±0.5 meters.
    \item Robust UAV Localization: The implementation of a 15-state EKF fusing GPS and IMU data to ensure stable aerial tracking.
    \item Adaptive Planning: The integration of the TEB (Timed Elastic Band) global planner to optimize trajectories based on real-time aerial maps.
\end{itemize}

These experimental results validate the thesis that heterogeneous collaboration is superior to single-agent approaches in unstructured, adverse environments. This work lays a solid foundation for the future deployment of autonomous robotic teams in some of the most challenging regions on Earth, offering a pathway to safer, more efficient operations in the face of extreme winter conditions.


\bibliographystyle{named}
\bibliography{references}

\clearpage
\appendix

\begin{table*}[htbp]
    \centering
    \renewcommand{\arraystretch}{1.4} 
    \setlength{\tabcolsep}{12pt}      
    
    \begin{tabular}{|>{\centering\arraybackslash}p{0.45\textwidth}|>{\centering\arraybackslash}p{0.45\textwidth}|}
        \hline
        \textbf{Component} & \textbf{Computation Overhead} \\ \hline
        Ardupilot & 0.090 cores \\ \hline
        Gazebo & 2 cores \\ \hline
        UAV Localization/RTabMap & 0.30 cores \\ \hline
        Segmentation & 4 cores \\ \hline
    \end{tabular}
    \caption{Computation Analysis (Phase-I)}
    \label{tab:placeholder}
\end{table*}

\begin{table*}[htbp]
    \centering
    \renewcommand{\arraystretch}{1.4} 
    \setlength{\tabcolsep}{12pt}      
    
    \begin{tabular}{|>{\centering\arraybackslash}p{0.45\textwidth}|>{\centering\arraybackslash}p{0.45\textwidth}|}
        \hline
        \textbf{Component} & \textbf{Computation Overhead} \\ \hline
        Ardupilot & 0.090 cores \\ \hline
        Gazebo & 2 cores \\ \hline
        UAV Localization/RTabMap & 0.50 cores \\ \hline
        Segmentation & 2 cores \\ \hline
    \end{tabular}
    \caption{Computation Analysis (Phase-II)}
    \label{tab:placeholder}
\end{table*}


\begin{table*}[htbp]
    \centering
    \renewcommand{\arraystretch}{1.4} 
    \setlength{\tabcolsep}{12pt}      
    
    \begin{tabular}{|>{\centering\arraybackslash}p{0.45\textwidth}|>{\centering\arraybackslash}p{0.45\textwidth}|}
        \hline
        Operating System & Ubuntu (20.04)\\ \hline
        System Processor & Intel(R) Core(TM) i5-11300H Processor \@ 3.1GHz\\ \hline
         Gazebo & 2 cores \\ \hline
        \end{tabular}
    \caption{Hardware specifications}
    \label{tab:placeholder}
\end{table*}

\input{Appendix/Algo1}




\end{document}

%% file: Appendix/Algo1.tex
\begin{algorithm*}[htbp]
\caption{Depth-Based UGV 3D Pose \& Orientation Extraction}
\label{alg:pose_extraction}
\begin{algorithmic}[1]
\REQUIRE $I_{RGB}$ (RGB image), $I_D$ (Depth image), $K$ (Camera intrinsic matrix), $M_{YOLO}$ (Trained YOLOv5 model)
\ENSURE $(X, Y, Z)$ (3D spatial coordinates), $\theta$ (Yaw/Heading angle)

\STATE Extract UGV bounding box using YOLOv5 
\[
[u_{min}, v_{min}, u_{max}, v_{max}] = M_{YOLO}(I_{RGB})
\]
\STATE Compute bounding box centroid $(u, v)$ and dimensions $(l_x, l_y)$ 
\[
u = \frac{u_{min} + u_{max}}{2}, \quad v = \frac{v_{min} + v_{max}}{2}
\]
\[
l_x = u_{max} - u_{min}, \quad l_y = v_{max} - v_{min}
\]
\STATE Extract UGV depth value $Z$ at the centroid 
\[
Z = I_D(v, u)
\]
\STATE Compute 3D coordinates using pinhole parameters $(f, c_x, c_y)$ from $K$ 
\[
X = -(u - c_x) \cdot \frac{Z}{f}, \quad Y = (c_y - v) \cdot \frac{Z}{f}
\]
\STATE Crop depth region of interest ($ROI_D$) around bounding box 
\[
ROI_D \gets I_D \left[ \left(v - \frac{l_y}{2}\right) : \left(v + \frac{l_y}{2}\right), \left(u - \frac{l_x}{2}\right) : \left(u + \frac{l_x}{2}\right) \right]
\]
\STATE Normalize $ROI_D$ to $[0, 1]$ 
\STATE Initialize binary mask $M$ of same shape as $ROI_D$
\FOR{each pixel $p \in ROI_D$}
    \IF{$0 < p \leq 0.5$}
        \STATE $M(p) \gets 255$ 
    \ELSE
        \STATE $M(p) \gets 0$ 
    \ENDIF
\ENDFOR
\STATE Apply morphological erosion and dilation on $M$ 
\STATE Extract contours $C = \{c_1, c_2, \dots, c_n\}$ from $M$ 
\STATE Select largest contour $c_{max}$ based on enclosed area 
\STATE Compute orientation $\theta$ via PCA / Ellipse fitting on $c_{max}$ 
\[
\theta = \operatorname{atan2}(\text{eigenvector}_{0,1}, \text{eigenvector}_{0,0}) 
\]
\RETURN $(X, Y, Z), \theta$
\end{algorithmic}
\end{algorithm*}